\documentclass[11pt,letterpaper]{article}
\usepackage{acl2014}
\usepackage{times}
\usepackage{latexsym}
\usepackage{epsfig}
\usepackage{hhline}
\usepackage[normalem]{ulem}
\usepackage{multirow}
\usepackage{amssymb}
\usepackage{rotating}
\usepackage{graphicx}
\usepackage{caption}
\usepackage{subcaption}
\usepackage{tikz-dependency}
\usepackage{url}
\usepackage{alltt}
\usepackage{graphicx}
\usepackage{color}
\usepackage{colortbl}
\usepackage{titlesec}
\usepackage{verbatim}

\definecolor{grey}{RGB}{225,225,225}
\definecolor{black}{RGB}{105,105,105}
\newcommand{\X}{\cellcolor{black}}

\title{Static and Dynamic Feature Selection in Morphosyntactic Analyzers}

\author{Bernd Bohnet$^{\spadesuit}$ ~ Miguel Ballesteros$^{\diamondsuit\clubsuit}$ ~ Ryan McDonald$^{\spadesuit}$~ Joakim Nivre$^{\heartsuit}$\\
$^{\spadesuit}$Google Inc. London, United Kingdom.\\
$^{\diamondsuit}$NLP Group, Pompeu Fabra University, Barcelona, Spain \\
$^{\clubsuit}$School of Computer Science, Carnegie Mellon University, Pittsburgh, PA, USA \\
$^{\heartsuit}$Uppsala University. Department of Linguistics and Philology. Uppsala, Sweden\\
{\small \tt \{bohnetbd,ryanmcd\}@google.com, miguel.ballesteros@upf.edu, joakim.nivre@lingfil.uu.se}
}

\date{}

\begin{document}
\maketitle
\begin{abstract}
We study the use of greedy feature selection methods for 
morphosyntactic tagging under a number of different conditions. 
We compare a static ordering of features to a dynamic ordering
based on mutual information statistics, and we apply the techniques
to standalone taggers as well as joint systems for tagging and parsing.
Experiments on five languages show that feature selection can result in 
more compact models as well as higher accuracy under all conditions, 
but also that a dynamic ordering works better than a static ordering and 
that joint systems benefit more than standalone taggers. We also show that
the same techniques can be used to select which morphosyntactic
categories to predict in order to maximize syntactic accuracy in a joint system.
Our final results represent a substantial improvement of the state of the art
for several languages, while at the same time reducing both the number of 
features and the running time by up to 80\% in some cases.
\end{abstract}

\section{Introduction}

Morphosyntactic tagging, whether limited to basic part-of-speech tags or using rich morphosyntactic features, is a fundamental task
in natural language processing, used in a variety of applications from machine translation \cite{habash2006} to information extraction \cite{banko2007}.
In addition, tagging can be the first step of a syntactic analysis, providing a shallow, non-hierarchical representation of syntactic structure.

Morphosyntactic taggers tend to belong to one of two different paradigms: \emph{standalone taggers} or \emph{joint taggers}.
Standalone taggers use narrow contextual representations, typically an $n$-gram window of fixed size. To achieve state-of-the-art 
results, they employ sophisticated optimization techniques in combination with rich feature representations \cite{brants00anlp,toutanova00,gimenez04,muller2013}.
Joint taggers, on the other hand, combine morphosyntactic tagging with deeper syntactic processing. The most common case is parsers that predict constituency structures jointly with part-of-speech tags \cite{charniak05,petrov06} or richer word morphology \newcite{goldberg08}.

In dependency parsing, pipeline models have traditionally been the norm, but recent studies have shown that joint tagging and dependency parsing can improve accuracy of both \cite{lee11,hatori11,bohnet12emnlp,tacl-bbjn}. Unfortunately, joint models typically increase the search space, making them more cumbersome than their pipeline equivalents. For instance, in the joint morphosyntactic transition-based parser of \newcite{tacl-bbjn}, the number of parser actions increases linearly by the size of the part-of-speech and/or morphological label sets. For some languages this can be quite large. For example, \newcite{muller2013} report morphological tag sets of size 1,000 or more.

The promise of joint tagging and parsing is that by trading-off surface morphosyntactic predictions with longer distance dependency predictions, accuracy can be improved. However, it is unlikely that every decision will benefit from this trade-off. Local $n$-gram context is sufficient for many tagging decisions, and parsing decisions likely only benefit from morphological attributes that correlate with syntactic functions, like case, or those that constrain agreement, like gender or number. At the same time, while standalone morphosyntactic taggers require large feature sets in order to make accurate predictions, it may be the case that fewer features are needed in a joint model, where these predictions are made in tandem with dependency decisions of larger scope. This naturally raises the question as to whether we can advantageously optimize feature sets at the tagger and parser levels in joint parsing systems to alleviate their inherent complexity.

We investigate this question in the context of the joint morphosyntactic parser of \newcite{tacl-bbjn}, focusing on optimizing and compressing feature sets via greedy feature selection techniques, and explicitly contrasting joint systems with standalone taggers. The main findings emerging from our investigations are:
\begin{itemize}
	\item Feature selection works for standalone taggers but is more effective in a joint system. This holds for model size as well as tagging accuracy (and parsing accuracy as a result).
	\item Dynamic feature selection strategies that take feature redundancy into account often lead to more compact models than static selection strategies with little loss in accuracy.
	\item Similar selection techniques can also reduce the set of morphological attributes to be predicted jointly with parsing, reducing the size of the output space at no cost in accuracy.
\end{itemize}
\noindent
The key to all our findings is that these techniques simultaneously compress model size and/or decrease the search space while increasing the underlying accuracy of tagging and parsing, even surpassing the state of the art in a variety of languages. With respect to the former, we observe empirical speed-ups upwards of 5x. With respect to the latter, we show that the resulting morphosyntactic taggers consistently beat state-of-the-art taggers across a number of languages.


\section{Related Work}
\label{sec:background}



Since morphosyntactc tagging interacts with other tasks such as word segmentation and syntactic parsing, there has been an increasing
interest in joint models that integrate tagging with these other tasks. This line of work includes joint tagging and word segmentation \cite{zhang08acl},
joint tagging and named entity recognition \cite{conf/tsd/MoraV12}, joint tagging and parsing \cite{lee11,li11,hatori11,bohnet12emnlp,tacl-bbjn}, and even 
joint word segmentation, tagging and parsing \cite{hatori12}. These studies often show improved accuracy from joint inference in one or all of
the tasks involved. 

Feature selection has been a staple of statistical NLP since its beginnings, notably selection via frequency cut-offs in part-of-speech tagging \cite{ratnaparkhi96}. Since then efforts have been made to tie feature selection with model optimization. For instance, \newcite{mccallum2003} used greedy forward selection with respect to model log-likelihood to select features for named entity recognition. Sparse priors, such as L1 regularization, are a common feature selection technique that trades off feature sparsity with the model's objective \cite{gao2007}. \newcite{martins2011} extended such sparse regularization techniques to allow a model to deselect entire feature templates, potentially saving entire blocks of feature extraction computation. However, current systems still tend to employ millions of features without selection, relying primarily on model regularization to combat overfitting.
Selection of morphological attributes has been carried out previously in \newcite{ballesteros2013effective} and selection of features under similar constraints was carried out by \newcite{bb2014automatic}.

\section{Feature Selection}
\label{featselection}

The feature selection methods we investigate can all be viewed as greedy forward selection,
shown in Figure~\ref{forward}. 
This paradigm starts from an empty set and considers features one by one. In each iteration, a model is generated from a training set and tested on a development set relative to some accuracy metric of interest.
The feature under consideration is added if it increases this metric beyond some threshold and discarded otherwise.

This strategy is similar to the one implemented in MaltOptimizer \cite{BallesterosNivre2014}. It differs from classic forward feature selection \cite{dellapietra97} in that it does not test all features in parallel, but instead relies on an ordering of features as input. This is primarily for efficiency, as training models in parallel for a large number of feature templates is cumbersome.

The set of features, $F$, can be defined as fully instantiated input features, e.g., \emph{suffix=ing}, or as feature templates, e.g., \emph{prefix}, \emph{suffix}, \emph{form}, etc. Here we always focus on the latter. By reducing the number of feature templates, we are more likely to positively affect the runtime of feature extraction, as many computations can simply be removed. 

\subsection{Static Feature Ordering} 

Our feature selection algorithm assumes a given order of the features to be evaluated against the objective function. One simple strategy is for a human to provide a \emph{static} ordering on features, that is fixed for traversal. This means that we are testing feature templates in a predefined order and keeping those that improve the accuracy. Those that do not are discarded and never visited again. In Figure~\ref{forward}, this means that the Order($F$) function is fixed throughout the procedure. In our experiments, this fixed order is the same as in Table \ref{table:templates}.

\begin{figure}[t]
\scalebox{0.9}{ 
\begin{minipage}{.65\linewidth}
\begin{footnotesize}
\sffamily
\begin{tabular}{rl}
\hline
\multicolumn{2}{l}{Let $F = \{F_1, \ldots, F_n\}$ be the full set of features,} \\ 
\multicolumn{2}{l}{let $M(X)$ be the evaluation metric for feature set $X$,}\\
\multicolumn{2}{l}{let Order($F$) be an ordering function over a feature set $F$,}\\
\multicolumn{2}{l}{and let $\Delta$ be the threshold.}\\\\
1 & $B \leftarrow 0$\\
2 & $X \leftarrow \emptyset$\\
3 & \textbf{while} $|F| \neq 0$ \\
4 & \hspace{0.45cm} Order($F$) \\
5 & \hspace{0.45cm} \textbf{if} $M(X \cup \{F_0\}) + \Delta > B$ \textbf{then}\\
6 & \hspace{0.8cm} $B \leftarrow M(X \cup \{F_0\})$\\
7 & \hspace{0.8cm} $X \leftarrow X \cup \{F_0\}$\\
8 & \hspace{0.45cm} $F \leftarrow F - \{F_0\}$\\
9 & \textbf{return} $X$\\
\hline
\end{tabular}
\end{footnotesize}
\end{minipage}
}
\caption{Algorithm for forward feature selection.}
\label{forward}
\end{figure}

\subsection{Dynamic Feature Ordering}
\label{mutual}

In text categorization, static feature selection based on correlation statistics is a popular technique \cite{yang1997}. The typical strategy in such offline selectors is to rank each feature by its correlation to the output space, and to select the top K features. This strategy is often called \emph{max relevance}, since it 
aims to optimize the features 
based solely to their predictive power.


Unfortunately, the $n$ best 
features selected by these algorithms might not provide the best 
result \cite{Hanchuan2005}.
Redundancy among the features is the primary reason for this, and
Peng et al.\  develop the \emph{minimal redundancy maximal relevance (MRMR)} technique to address this problem.
The MRMR method tries to keep the redundancy minimal among the 
features. 
The approach is based on mutual information to compute the relevance of features
and the redundancy of a feature in relation to a set of already selected features.
 
The mutual information of two discrete random variables $X$ and $Y$ is defined as follows

\begin{center}
$I(X;Y)=\sum\limits_{x\in X} \sum\limits_{y\in Y} p(x,y)log_2\frac{p(x,y)}{p(x)p(y)}$
\end{center}
Max relevance selects the feature set $X$ that maximizes the mutual information of feature templates $X_i \in X$ and the output classes $c \in C$.
\[
\max~D(X,C), ~~D(X,C)=\frac{1}{|X|}\sum\limits_{X_i \in X} I(X_i;C).
\]
To account for cases when features are highly redundant, and thus would not change much the discriminative power of a classifier, the following criterion can be added to minimize mutual information between selected feature templates:
\[
\min~R(X), ~~R(X)=\frac{1}{|X^2|}\sum\limits_{X_i,X_j\in X} I(X_i;X_j)
\]
Minimal redundancy maximal relevance (MRMR) combines both objectives:
\[
\max~\Phi(D,R), ~~\Phi=D(X,C)-R(X)
\]
For the greedy feature selection method outlined Figure~\ref{forward}, we can use the MRMR criteria to define the Order($F$) function. This leads to a \emph{dynamic} feature selection technique as we update the order of features considered dynamically at each iteration, taking into account redundancy amongst already selected features.

This technique can be seen in the same light as greedy document summarization \cite{carbonell1998}, where sentences are selected for a summary if they are both relevant and minimally redundant with sentences previously selected. 

\section{Morphosyntactic Tagging}

In this section, we describe the two systems for morphosyntactic tagging we use to compare feature selection techniques.

\subsection{Standalone Tagger}

The first tagger is a standalone SVM tagger, whose training regime is shown in
Figure~\ref{tagger}. The tagger iterates up to $k$ times (typically twice) over a sentence from left to right (line 2).  This iteration is performed to allow the final assignment of tags to benefit from tag features on both sides of the target token.
For each token of the sentence, the tagger initializes an $n$-best list and extracts features for the token in question (line 4-7).
In the innermost loop (line 9-11), the algorithm computes the score for each morphosyntactic tag and inserts  a pair consisting of the morphosyntactic tag and its score into the $n$-best list. The algorithm returns a two dimensional array, where the first dimension contains the tokens and the second dimension contains the sorted lists of tag-score pairs. The tagger is trained online using MIRA \cite{crammer06}. When evaluating this system as a standalone tagger, we select the 1-best tag. 
This can be viewed as a multi-pass version of standard SVM-based tagging \cite{marquez2004}. 


\begin{figure}[t]
\scalebox{0.9}{ 
\begin{minipage}{.65\linewidth}
\begin{footnotesize}
\sffamily
\begin{tabular}{rl}
\hline
\multicolumn{2}{l}{$x \in X $ is a sentence and $w_{dim(P)}$ are weight vectors} \\ 
\multicolumn{2}{l}{\sc{tag}$(x,w_{dim(P)})$}\\
1  & // iterate k times over each sentence\\
2 & \textbf{for} $j\leftarrow 1~ ..~ k$\\
3  & \hspace{0.45cm}// for each token in sentence $x$\\
4 &\hspace{0.45cm} \textbf{for} $t \leftarrow~1~..~length(x)$\\
5  & \hspace{0.80cm}  $nbest[t] \leftarrow []$\\
6  & \hspace{0.80cm} // extract the features for token $t$\\
7  & \hspace{0.80cm}  $f_t \leftarrow f(x,t,nbest)$\\
8  & \hspace{0.80cm} // for each part-of-speech tag $p$\\
9 & \hspace{0.80cm}  \textbf{for each} $p \in P$\\
10 & \hspace{1.20cm}  $score \leftarrow w_p \cdot f_t$ \\
11 & \hspace{1.20cm}  $nbest[t] \leftarrow $\sc{insert-pair} $(nbest[t], (p,score))$ \\
12 & \textbf{return} $nbest$\\
\hline
\end{tabular}
\end{footnotesize}
\end{minipage}
}
\caption{\label{tagger}Algorithm for standalone tagger}
\end{figure}

\subsection{Joint Dependency-Based Tagger}

The joint tagger-parser follows the design of \newcite{tacl-bbjn}, who augment an arc-standard transition-based dependency parser 
with the capability to select a part-of-speech tag and/or morphological tag for each input word from an $n$-best list of tags for that word. 
The tag selection is carried out when an input word is shifted onto the stack. Only the $k$ highest-scoring tokens from each $n$-best list are considered, and only tags 
whose score is at most $\alpha$ below the score of the best tag. In all experiments of this paper, we set $k$ to 2 and $\alpha$ to 0.25. 
The tagger-parser uses beam search to find the highest scoring combined tagging and dependency tree. When pruning the beam, it 
first extracts the 40 highest scoring distinct dependency trees and then up to 8 variants that differ only with respect to the tagging,
a technique that was found by \newcite{tacl-bbjn} to give a good balance between tagging and parsing ambiguity in the beam.
The tagger-parser is trained using the same online learning algorithm as the standalone tagger.
When evaluating this system as a part-of-speech tagger, we consider only the finally selected tag sequence in the dependency 
tree output by the parser. 


\section{Part-of-Speech Tagging Experiments}
\label{sec:experiments}

To simplify matters, we start by investigating feature selection for part-of-speech taggers, both in the context of standalone and joint systems. 
The main hypotheses we are testing is whether feature selection techniques are more powerful in joint morphosyntactic systems as opposed to standalone taggers. That is, the resulting models are both more compact and accurate. Additionally, we wish to empirically compare the impact of static versus dynamic feature selection techniques.

\subsection{Data Sets}
\label{exsetup}

We experiment with corpora from five different languages: Chinese, English, German, Hungarian and Russian. For Chinese, we use the Penn Chinese Treebank 5.1 (CTB5), converted with the head-finding rules, conversion tools and with the same split as in \newcite{zhang08}.\footnote{Training: 001--815, 1001--1136. Development: 886--931, 1148--1151. Test: 816--885, 1137--1147.} For English, we use the WSJ section of the Penn Treebank, converted with the head-finding rules of \newcite{yamada03} and the labeling rules of \newcite{nivre06book}.\footnote{Training: 02-21. Development: 24. Test: 23.} For German, we use the Tiger Treebank \cite{brants02} in the improved dependency conversion by \newcite{seeker12}. For Hungarian, we use the Szeged Dependency Treebank \cite{farkas12}. For Russian we use the SynTagRus Treebank \cite{boguslavsky00,boguslavsky02}. 

\subsection{Feature Templates}

Table~\ref{table:templates} presents the
feature templates that we employed in our experiments (second column). 
The name of the functor indicates the purpose of the feature template.
For instance, the functor {\em form} defines the word form. 
The argument specifies the location 
of the token, for instance, {\em form(w+1)} denotes 
the token to the right of the current token {\em w}. 

When more than one argument is given, the functor is applied 
to each defined position and the results are concatenated. Thus,
 {\em form(w,w+1)} expands to {\em form(w)+form(w+1)}.
The functor \emph{formlc} denotes the form with all letters converted to lowercase
and \emph{lem} denotes the lemma of a word. 
The functors {\em suffix1, suffix2,...} and {\em prefix1,...} denote
suffixes and prefixes of length {\em 1, 2, ..., 5}.
The {\em suffix1+uc, ...} functors concatenates a suffix with a value that indicates
uppercase or lowercase word. 
 
The functors {\em pos} and {\em mor} denote part-of-speech tags and morphological tags, respectively.
The tags to the right of the current position are available as features in the second iteration of the standalone tagger 
as well as in the final resolution stage in the joint system. 
Patterns of the form $c_i$ denote the $i$th character.  Finally, the functor {\em number}
denotes a sequence of numbers, with optional periods and commas. 

\subsection{Main Results}
\label{fsa}

In our experiments we make a division of the training corpora into 80\% for training and 20\% for development. Therefore, in each iteration a model is trained over 80\% of the training corpus and tested on 20\%.\footnote{There is also a held-out test set for evaluation, which is the standard test set provided and depicted in Section \ref{exsetup}.} For feature selection, if the outcome of the newly trained model is better than the best result so far, then the feature is added to the feature model; otherwise, it is not. A model has to show improvement of at least 0.02 on part-of-speech tagging accuracy to count as better.%
\footnote{All the experiments were carried out on a CPU Intel Xeon 3.4 Ghz with 6 cores. Since the feature selection experiments require us to train a large number of parsing and/or tagging models, we needed to find a realistic training setup that gives us a sufficient accuracy level while maintaining a reasonable speed. 
After some preliminary experiments, we selected a beam size of 8 and 12 training iterations for the feature selection experiments while the final models are tested with a beam size of 40 and 25 training iterations. The size $k$ of the second beam for alternative tag sequences is kept at 8 for all experiments and the threshold $\alpha$ at 0.25.}

\begin{table*}
\centering
\renewcommand{\tabcolsep}{2pt} 
\renewcommand{\arraystretch}{1}
\scalebox{0.9}{ 
\begin{scriptsize}
\begin{tabular}{|l|l|c|c|c|c|c||c|c|c|c|c||c|c|c|c|c||c|c|c|c|c||c|c|c||c|c|c||c|c|c||c|c|c|c|c|c}
\hline
 & &\multicolumn{20}{|c|}{{\bf Part-of-Speech}} &\multicolumn{12}{|c|}{{\bf Morphology}} \\\hline 
 & &\multicolumn{10}{|c|}{{\bf Standalone}} &\multicolumn{10}{|c|}{{\bf Joint}} &\multicolumn{6}{|c|}{{\bf Standalone}} &\multicolumn{6}{|c|}{{\bf Joint}} \\\hline 
 & &\multicolumn{5}{|c|}{{\bf static}}&\multicolumn{5}{|c|}{{\bf dynamic}}&\multicolumn{5}{|c|}{{\bf static}} &\multicolumn{5}{|c|}{{\bf dynamic}} &\multicolumn{3}{|c|}{{\bf static}}&\multicolumn{3}{|c|}{{\bf dynamic}}&\multicolumn{3}{|c|}{{\bf static}}&\multicolumn{3}{|c|}{{\bf dynamic}} \\\hline
Id& Feature Name &Ch & En & Ge & Hu & Ru & Ch & En & Ge & Hu & Ru & Ch & En & Ge & Hu & Ru& Ch & En & Ge & Hu & Ru & Ge & Hu & Ru & Ge & Hu & Ru & Ge & Hu & Ru & Ge & Hu & Ru  \\\hline
1 & form         &\X & \X & \X &\X  &\X  & \X & \X &    &    &    & \X & \X &\X  &\X  &\X & \X &    &    &   &      &\X  &\X &\X  &    &    &\X    &\X    &\X    &\X      &  &   &  \\\hline
2 & formlc       &\X &    &    &\X  &\X    &    &  \X   &\X  &    &    &    &    &    &\X  &\X   & \X &  \X  &\X  &\X    &   &    &\X    &\X    &\X    &\X    &    &    &\X    &\X    &   &\X    &  \\\hline
3 & prefix1      &   &  \X & \X &\X    &\X    & \X &    &    &\X    &\X    & \X & \X   &\X    &\X    &\X   &\X &\X & &\X      &   &\X    &\X    &\X    &    &    &    &\X    &\X    &\X   &    &  & \\\hline
4 & prefix2      & & \X & &\X &\X & & &\X & && &  \X  &    &    &\X  &  &  &\X  &  &  &    &\X    &    &    &    &    &    &\X    &\X    &    &    & \\\hline
5 & prefix3      & & & &\X &\X & & \X &\X &\X && &  &  &\X  &\X  &  &  &  &  &&    &    &    &    &    &    &    &\X    &    &    &    &       \\\hline
6 & prefix4      &   &    & \X &\X    &\X    &    & &\X & && &  &  &  &\X  &  &  &\X  &  &&    &    &    &    &    &    &    &    &    &    &    &      \\\hline
7 & prefix5      &   &  & &\X &\X & & & & & & &  &  &  &\X  &  &  &  &  & &  &  &  &  &  &  &  &  &  &  &  &\\\hline
13& suffix1      &\X & \X  & \X  &\X   &\X    & \X &\X  &\X  &\X &\X         &\X & \X   &\X    &\X    &\X   &\X & &\X &\X &\X  &\X  &\X  &\X  &\X  &\X  &\X  &\X &\X  &\X  &\X  &\X  &\X \\\hline
14& suffix2      &   & \X  & \X &\X    &\X    &   &\X  &\X  &\X &    &   &  \X  &\X    &\X    &\X   &\X  & & &\X &\X  &\X  &\X  &\X  &  &  &\X  &\X  &\X  &\X   &\X  &  &\X\\\hline
15& suffix3      &   & \X  & \X &\X &\X & & & &\X &\X& & \X&\X &\X &\X & & & &\X &\X&\X  &\X  &\X  &\X  &  &\X  &\X  &\X  &\X   &  &\X  &\X\\\hline
16& suffix4      &   & \X  & \X &\X   &\X      & \X &    &    &   &\X     &\X & &\X &\X &\X & & & & &&\X  &\X  &\X  &  &\X  &\X  &\X  &\X  &\X   &\X  &  &\X\\\hline
17& suffix5      &   &   & \X &\X   &\X      &    &    &    &   &     &   &\X &\X &\X & & &\X & & &&\X  &\X  &\X  &  &  &  &\X  &  &\X   &\X  &  &\\\hline
18 & suffix2+uc  &    & & &\X & & & & & &      &   &   &\X    &     &   &\X &\X & & &&  &\X  &  &\X  &\X  &\X  &  &  &\X   &\X  &\X  &\\\hline
19 & suffix3+uc  &    & & & & & & &\X & &      &   &   &    &     &   & & & & &&\X  &  &  &  &\X  &  &  &  &   &\X  &\X  &\\\hline
20 & suffix4+uc  &    & & & & & & \X & & &      &   &   &    &     &   &\X &\X &\X & &\X&  &  &  &\X  &  &\X  &  &  &   &\X  &  &\\\hline
21 & suffix5+uc  &    & & & & & &\X  &\X &\X &      &   &   &    &     &\X   &\X & &\X &\X &&  &  &  &\X  &\X  &\X  &  &  &   &  &\X  &\X\\\hline
22 & uppercase   & & & & & & &\X  &\X & &\X&        &    &\X    &   &   & \X &\X &\X & &&\X  &  &  &\X  &\X  &\X  &\X  &\X  &\X   &\X  &\X  &\X\\\hline
23 &c$_2$c$_3$c$_4$&\X& \X  &   &   &   & \X & & & &\X   &    &    &\X    &   &   & \X& & & &&  &  &  &  &  &  &  &\X  &   &\X  &  &\\\hline
24 &c$_3$c$_4$c$_5$&  & \X& & && & & & &       &\X  &    &    &   &   & \X& & & &&  &  &  &  &  &  &  &  &   &  &  &\X\\\hline
25 &c$_4$c$_5$c$_6$&\X& & & & & & & & &      &    &    &    &   &\X   & \X& & & &&  &\X  &  &  &  &  &  &  &   &\X\X  &  &\\\hline
26 & c$_2$c$_3$c$_4$c$_5$&&&\X& & & & & && & & \X& & & & & & & &&  &  &  &\X\X  &\X  &\X\X  &  &\X  &   &\X  &  &\\\hline
27 & c$_3$c$_4$c$_5$c$_6$&&&\X& &\X & & &\X &\X&\X & & &\X & & & & &\X & &&  &  &  &  &  &  &  &  &   &  &  &\\\hline
28 &form(w,w+1) &\X & \X  & \X  &\X    &\X    & \X & \X   & & &   &\X &\X &\X &\X &\X & & &\X &\X &&\X  &\X  &\X  &  &  &  &\X  &\X  &\X   &  &  &\\\hline
29 & form(w+1)   &\X & \X  & \X  &\X    &\X    & \X & &\X &\X &\X     &\X & &\X & & & & & &\X &&\X  &\X  &\X  &\X  &\X  &  &\X  &  &   &\X  &\X  &\\\hline
30 & prefix1(w+1)&\X &   & \X  &\X    &    & & &\X &\X &   &    &    &\X   &   & &\X & &\X & &&\X  &\X  &  &  &  &  &\X  &  &   &  &  &\\\hline
31 & suffix1(w+1)&   &  \X & \X  &    &\X    & & \X &\X &\X &   &    &  \X  &   &   & & & &\X &\X &&\X  &  &\X  &  &  &  &\X  &  &\X   &\X  &  &\\\hline
32 & suff2+pref1(w+1)& & & & &  & \X & & &\X &         &\X & & & & & & & & &&  &  &  &  &  &  &  &  &   &  &  &\\\hline
33 & suff2+suff1(w+1)& & & & &\X & & & & &&       &    &    &   &\X   & \X & & & &&  &  &  &  &  &  &  &  &   &  &  &\\\hline
34 & suffix2(w+1)        & & \X & & & & & \X & &\X &&       &    &    &   &   & \X & & & &&\X  &  &  &  &  &  &\X  &  &   &  &  &\\\hline
35 & prefix2(w+1)        & &  & & & & &  & & &&       &    &    &   &   &  & & & &&  &  &\X  &  &  &  &\X &  &\X   &\X  &  &\\\hline
36 & suff2+pref2(w+1)&& & & && &\X  &\X & &  & & & & & & & &\X & &&  &  &  &  &  &  &  &  &   &  &  &\\\hline
37 & suffix2(w,xw+1)&& & & && & & & &\X  & & & & &\X & & & & &&  &  &  &  &  &\X  &  &\X  &   &\X  &  &\\\hline
41 & form(w+1,w+2)&\X &   & \X &\X   &   & \X & & & &    &\X &\X & & & & & & & &&\X  &  &\X  &  &  &  &\X  &  &   &  &  &\\\hline
42 & form(w+2)&   &   &    &     &   &    & & & &    &   &   & & & & & & & &&\X    &  &\X  &\X  &  &\X  &\X  &  &   &\X  &  &\\\hline
43 & form(w+2,w+3)  &  & \X &  &   &    &  & & &\X &\X  & & \X &\X &\X   & & & & & &&\X  &  &\X  &  &  &  &\X  &  &   &  &  &\X\\\hline
44 & length(w)      &  &   & \X &   &   & \X &\X  &  &\X&      &\X  & &\X     &   &\X   & \X& \X&\X &\X &&\X  &  &\X  &  &  &  &  &  &   &\X  &\X  &\\\hline
45 & lemma(w) &  &   &   &   &   & \X &\X  & &\X&\X        &\X & & & & & & &\X &\X &&  &  &  &  &\X  &  &  &\X  &   &  &\X  &\\\hline
46 & number   &\X&   &   &\X   &\X   & \X & &\X &\X &\X      &\X & & &\X &\X & & &\X &\X &\X&  &\X  &\X  &\X  &  &\X  &  &\X  &\X   &\X  &  &\X\\\hline
47 &lemmas(w-1,w+1)&\X& \X & \X &\X   &\X    & \X & & & &      &\X & \X &\X &\X & & & & & &\X&\X  &\X  &\X  &  &  &  &\X  &  &\X   &  &  &\\\hline
48 & form(w-1)     &\X& \X & \X  &\X   &\X    & \X & & & &      &\X &  \X &\X   &\X   &\X    &\X& & &\X &&\X  &\X  &\X  &\X  &  &\X  &\X  &\X  &\X   &\X  &  &\\\hline
49 & lemma(w-1 )   &  & \X & \X &   &\X    &  &\X  &\X &\X &\X      &  & & & & & & &\X & &&\X  &  &\X  &\X  &  &\X  &  &  &\X   &\X  &  &\X\\\hline
50 & form(w-2)     &\X& \X & \X &   &\X    & \X & & & &    &\X & &\X & & & & & & &&\X  &  &\X  &\X  &  &  &\X  &  &   &\X  &  &\X\\\hline
51 & form(w-3,w-2)  & &\X  &  &   &    &  & & &\X &    & &\X & & & & & & & &&\X  &\X  &  &  &  &  &\X  &  &   &  &  &\\\hline
52 & form(w-1,w)  &  &  & \X &\X   &    & \X & \X & &\X &       &\X & & & & & &\X & & &&\X  &\X  &\X\X  &  &  &  &\X  &  &   &\X  &  &\\\hline
53 & form(w-2,w-1,w)  &  &  &  &   &    &  & & & &\X       & & & & & & & &\X & &&\X  &  &  &  &  &  &  &  &   &  &  &\\\hline
54 & form(w-1,w,w+1)  &  &  &  &   &    &  & & & &       & & & & & & & & & &\X&  &  &  &  &  &  &  &  &   &  &  &\\\hline
55 & form(w,w+1,w+2)  &  &  &  &   &    &  & & & &       & & & & & & & &\X & &&\X  &  &  &  &  &  &  &  &   &  &  &\\\hline
56 & suffix2(w-1)  &  &  &  &\X   &    &    & & &\X &          &          &    &    &    &    &  & & &\X &&\X  &  &\X  &\X  &  &\X  &\X  &  &\X   &\X  &  &\X\\\hline
57 & suffix2(w-1,w-2)  &  &  &  &   &    &    &\X & &\X &          &          &    &    &    &    &  & & &\X &&\X  &\X  &\X  &  &  &  &  &  &\X   &  &  &\\\hline
59 & suffix2(w-2)  &  &  &  &   &    &    & & & &          &          & \X   &    &    &    &  & & & &&\X  &  &  &\X  &  &\X  &  &  &   &\X  &  &\\\hline
60 & suffix2(w-3)  &  &  &    &   &    &    & & & &          &          &    &    &      &    &  & & &   &&    &  &    &\X  &  &  &  &  &   &  &  &\\\hline
61 & suffix2(w+1)  &  &  & \X &   &    &    & & & &          &          &    &    &\X    &    &  & & &\X &&\X  &  &\X  &  &  &  &  &  &   &  &  &\\\hline
64 & suffix2(w+2)  &  &  &    &   &    &    & & & &          &          &    &    &      &    &  & & &   &&    &  &  &  &  &  &\X  &  &   &  &  &\\\hline
67 & pos(w+1,w+2)&\X&\X&\X & & & &\X  & &\X &    &          &    &    &    &    & \X & & & &&  &  &  &  &  &  &  &  &   &  &  &\\\hline
68 & pos(w+1)      &\X& \X & \X &\X    &\X & & \X &\X &\X &\X & & & & & & &\X &\X &\X &&  &\X  &  &  &  &  &  &  &   &  &  &\X\\\hline %
69 & pos(w-1)      &\X& \X & \X &    &    & \X &  \X  &\X &\X &\X          &\X  &   &   &    &\X    &\X&\X &\X &\X &\X&\X  &  &\X  &\X  &  &  &  &  &   &\X  &  &\\\hline
70 & pos(w-1,w+1)&\X &\X& & & & & \X &\X & &\X& & & & & & & & & &\X&  &  &  &  &  &  &  &  &   &\X  &  &\\\hline
71 & pos(w-2)&\X&   &   &   &   & \X &  \X  &\X & &         &\X & & & & & &\X &\X &\X &&  &  &  &  &  &\X  &  &  &   &  &  &\\\hline
72 & pos(w-1,w-2)&&&&  &   & \X & & & &      &\X &   &   &    &    &\X& & & &&  &  &  &  &  &  &  &  &   &\X  &  &\\\hline
73 & form(w-1,w-2)&&&&  &   &  & & &\X &      & &   &   &    &    & & & & &&  &  &  &\X  &  &\X  &  &  &   &  &  &\\\hline
74 & form(w-1,w-1)&&&&  &   &  & & & &      & &   &   &    &    & & \X& & &&  &  &  &  &  &  &  &  &\X   &  &  &\\\hline
75 & pos(w-3)& &   &   &   &   &  &   &\X & &         & & & & & & & & & &&  &  &  &  &  &  &  &  &   &  &  &\\\hline

76 &morph(w+1,w+2)& &   &   &   &   &  &   & & &         & & & & & & & & & &&\X  &  &\X  &\X  &  &  &\X  &  &   &  &  &\\\hline
77 &morph(w+1)& &   &   &   &   &  &   & & &         & & & & & & & & & &&\X  &  &\X  &\X  &\X  &\X  &\X  &  &   &  &  &\\\hline
78 &morph(w-1)& &   &   &   &   &  &   & & &         & & & & & & & & & &&\X  &\X  &\X  &  &\X  &\X  &  &\X  &   &  &\X  &\X\\\hline
79 &morph(w-1,w+1)& &   &   &   &   &  &   & & &         & & & & & & & & & &&\X  &  &  &  &  &  &  &  &   &  &  &\\\hline
80 &morph(w-2)& &   &   &   &   &  &   & & &         & & & & & & & & & &&\X  &  &\X  &  &  &  &\X  &  &   &\X  &  &\\\hline
81 &morph(w-1,w-2)& &   &   &   &   &  &   & & &         & & & & & & & & & &&\X  &  &  &\X  &  &  &  &  &   &  &  &\\\hline
82 &fo.(w-1)+mo.(w-2)& &   &   &   &   &  &   & & &         & & & & & & & & & &&\X  &  &  &  &  &\X  &\X  &  &   &\X  &  &\\\hline
83 &morph(w-2,w-3)& &   &   &   &   &  &   & & &         & & & & & & & & & &&\X  &  &  &\X  &  &  &\X  &  &   &  &  &\\\hline
84 &pos(w)& &   &   &   &   &  &   & & &         & & & & & & & & & &&\X  &\X  &\X  &\X  &\X  &\X  &  &  &   &\X  &\X  &\X\\\hline
85 &pos(w,w-1)& &   &   &   &   &  &   & & &         & & & & & & & & & &&\X  &  &  &\X  &\X  &\X  &  &  &   &  &  &\X\\\hline
86 &pos(w,w+1)& &   &   &   &   &  &   & & &         & & & & & & & & & &&\X  &  &  &\X  &\X  &  &  &  &   &\X  &  &\\\hline
87 &pos(w-2,w-1,w)& &   &   &   &   &  &   & & &         & & & & & & & & & &&\X  &  &  &\X  &  &  &  &  &   &\X  &  &\\\hline
88 &pos(w,w+1,w+2)& &   &   &   &   &  &   & & &         & & & & & & & & & &&\X  &  &  &  &  &  &  &  &   &  &  &\\\hline
89 &pos(w-1,w,w+1)& &   &   &   &   &  &   & & &         & & & & & & & & & &&  &\X  &  &  &  &  &  &  &   &  &  &\\\hline
\# & total &21& 24 &23 &23 &23 & 19 &24 &21&24 &17   &19 &16  &19   &15   &21    &20& 15 &21&18 &9&44  &24  &31  &29  &15&24  &27 &18  &21   &32  & 12  &16\\\hline
\# & average &\multicolumn{5}{|c|}{ 22.8} &\multicolumn{5}{|c|}{ 21}  & \multicolumn{5}{|c|}{17.8} & \multicolumn{5}{|c|}{16.6} & \multicolumn{3}{|c|}{31}  & \multicolumn{3}{|c|}{22.6}   & \multicolumn{3}{|c|}{22}& \multicolumn{3}{|c|}{20}\\\hline
 & reduction & \multicolumn{5}{|c|}{69\%}  & \multicolumn{5}{|c|}{ 71\%}  & \multicolumn{5}{|c|}{ 76\%} & \multicolumn{5}{|c|}{78\%} & \multicolumn{3}{|c|}{58\%}  & \multicolumn{3}{|c|}{69\%}   & \multicolumn{3}{|c|}{70\%}& \multicolumn{3}{|c|}{73\%}\\\hline
\end{tabular}
\end{scriptsize}
}
\caption{Selected features for the standalone tagger and the joint tagger-parser with a threshold of 0.02 for the experiments of Section \ref{sec:experiments} (part-of-speech tagging) and Section \ref{morphology} (morphological tagging). Some of the features are not shown because they have not been selected. A filled box means selected, while an empty one means not selected. }
\label{table:templates}
\end{table*}

Table \ref{table:templates} (columns under Part-of-Speech) shows the features that the algorithms selected for each language and each system, and Table \ref{table:pos-results-dev} shows the performance on the development set. We primarily report part-of-speech tagging accuracy (POS), but also report unlabeled (UAS) and labeled (LAS) attachment scores \cite{buchholz06} to show the effect of improved taggers on parsing quality. Additionally, Table~\ref{table:pos-results-dev} contains the number of features selected (\#).

The first conclusion to draw is that the feature selection algorithms work for both standalone and joint systems.
The number of features selected is drastically reduced. The dynamic MRMR feature selection technique for the joint system compresses the model by as much as 78\%. This implies faster inference (smaller dot products and less feature extraction) and a smaller memory footprint. In general, joint systems compress more over their standalone counterpart, by about 20\%. Furthermore, the dynamic technique tends to have slightly more compression.

The accuracies of the joint tagger-parser are in general superior to the ones obtained by the standalone tagger,
as noted by \newcite{bohnet12emnlp}.
In terms of tagging accuracy, static selection works slightly better for Chinese, German and Hungarian while dynamic MRMR works best for English and Russian (Table~\ref{table:pos-results-dev}).
Moreover, the standalone tagger selects several feature templates that requires iterating over the sentence, such as {\em\small pos(w+1), pos(w+2)}, whereas the feature templates selected by the joint system contain significantly fewer of these features. 
This shows that a joint system is less reliant on context features to resolve many ambiguities that previously needed a wider context. This is almost certainly due to the global contextual information pushed to the tagger via parsing decisions.
As a consequence, the preprocessing tagger can be simplified and we need to conduct only one iteration over the sentence while maintaining the same accuracy level. 
Interestingly, the dynamic MRMR technique tends to select less \emph{form} features, which have the largest number of realizable values and thus model parameters. 

Table~\ref{table:tagger-state-of-art} compares the performance of our two taggers with two state-of-the-art taggers. Except for English, the joint tagger consistently outperforms the Stanford tagger and MarMot: for Chinese by 0.3, for German by 0.38, for Hungarian by 0.25 and for Russian by 0.75. Table~\ref{table002} compares the resulting parsing accuracies to state-of-the-art dependency parsers for English and Chinese, showing that the results are in line with or higher than the state of the art.

\begin{table}[t]
\small
\centering
\renewcommand{\tabcolsep}{1.0pt} 
\scalebox{1.0}{ 
\begin{tabular}{|r|r|r|r|r||l|c|c|c|c|c|c|c|c|c|c|c|c|c|c|c|c|c|c|c|c|c|c}
\hline
          &\multicolumn{4}{|c|}{{\bf Pipeline}}&\multicolumn{4}{|c|}{{\bf Joint}} \\\hline
          & POS   & LAS   & UAS   & \#       & POS   & LAS   & UAS    & \# \\\hline
  \multicolumn{9}{|c|}{{\bf Chinese}} \\\hline
none      & 94.14 & 78.98 & 81.77 & 74       & 94.34 & 79.33 & 82.21  & 74 \\\hline
static    & 94.26 & 79.06 & 81.96 & 20       &\bf 94.57 & 79.75 & 82.54  & 19 \\\hline
dynamic   & 94.06 & 79.26 & 82.21 & 23       & 94.49& 79.68& 82.54& 20 \\\hline
  \multicolumn{9}{|c|}{{\bf English}} \\\hline
none      & 97.18 & 90.78 & 91.97 & 74       & 96.99 & 90.92 & 92.06  & 74 \\\hline
static    & 96.98 & 90.56 & 91.77 & 23       & 96.99 &  91.05 &  92.20  & 16 \\\hline
dynamic   & 97.05 & 90.69 & 91.97 & 23       & \bf 97.13 & 90.78 & 91.95  &  15\\\hline 
  \multicolumn{9}{|c|}{{\bf German}} \\\hline
none      & 97.79 & 91.33 & 93.33 & 74       & 98.14 & 91.64 & 93.60  & 74 \\\hline 
static    & 97.77 & 91.70 & 93.71 & 25       & \bf 98.20 & 91.83 & 93.83  & 19 \\\hline 
dynamic   & 97.60 & 91.40 & 93.60 & 22       & 97.91 & 91.74 & 93.73  & 22 \\\hline 
  \multicolumn{9}{|c|}{{\bf Hungarian}} \\\hline
none      & 97.89 & 87.89 & 90.44 & 74       & 98.00 & 88.11 & 90.59  & 74 \\\hline 
static    & 97.87 & 88.01 & 90.54 & 23       & \bf 98.01 &  88.25 &  90.78  & 23 \\\hline 
dynamic   & 97.85 & 87.80 & 90.36 & 24       & 98.00 & 88.16 & 90.65  &  18 \\\hline 
  \multicolumn{9}{|c|}{{\bf Russian}} \\\hline
none      & 98.62 & 87.45 & 92.58 & 74       & 98.79 & 87.69 & 92.83  & 74 \\\hline 
static    & 98.70 & 87.41 & 92.62 & 23       & 98.85    &  87.61 &  92.86  & 21 \\\hline 
dynamic   & 98.69 & 87.69 & 92.92 & 17       & \bf 98.87 &  87.61 &  92.86 & 9 \\\hline 
\end{tabular}
}
\caption{Tagging and parsing accuracy scores on the dev set without feature selection (none), with static and with dynamic MRMR greedy feature selection. 
}
\label{table:pos-results-dev}
\end{table}

\begin{table}[t]
\small
\centering
\renewcommand{\tabcolsep}{2pt} 
\begin{tabular}{|l|l|l|l|l|l|l|l|l|l|l|}
\hline
&\multicolumn{1}{|c|}{Ch} &\multicolumn{1}{|c|}{En} &\multicolumn{1}{|c|}{Ge} & \multicolumn{1}{|c|}{Hu} &\multicolumn{1}{|c|}{Ru}\\\hline
             & POS  & POS     & POS  & POS & POS  \\\hline
Stanford     & 93.75 &\bf 97.44 & 97.51 & 97.55& 98.16 \\\hline
MarMot       & 93.84 & 97.43 & 97.57& 97.63 & 98.18  \\\hline
Standalone   & 94.04     & 97.33 & 97.56 & 97.69 & 98.73 \\\hline
Joint        & \bf 94.14 &  97.42  &\bf 97.95 & \bf 97.88  & \bf 98.93  \\\hline
\end{tabular}
\caption{State-of-the-art comparison for tagging on the test set.}\label{table:tagger-state-of-art}
\vspace{-0.3cm}
\end{table}

\begin{table}[t]
\small
\centering
\renewcommand{\tabcolsep}{1.0pt} 
\scalebox{1.0}{ 
\begin{tabular}{|r|r|r|r|r|l|c|c|c|c|c|c|c|c|c|c|c|c|c|c|c|c|c|c|c|c|c|c}
\hline
System    &            \multicolumn{4}{|c|}{} \\\hline
          &            POS   & LAS   & UAS    & \# \\\hline
  \multicolumn{5}{|c|}{{\bf Chinese}} \\\hline
\newcite{li11}            & 93.08 &      &80.55 & \\
\newcite{hatori12}        & 93.94 &      &81.20 & \\
\newcite{bohnet12emnlp}   & 93.24 & 77.91 &81.42 & \\\hline 
joint-static           & 94.14 & 78.77 & 81.91  & 20 \\\hline
  \multicolumn{5}{|c|}{{\bf English}} \\\hline
\newcite{mcdonald05acl}   &&    &90.9~~ &   \\
\newcite{mcdonald06eacl}  &&    &91.5~~ &   \\
\newcite{huang10}         &&    &92.1~~ &   \\
\newcite{koo10acl}        &&    &93.04&     \\
\newcite{zhang11}    &  &    &92.9~~ &   \\
\newcite{martins10}       &&    &93.26&\\ 
\newcite{bohnet12emnlp}   & 97.33 & 92.44 & 93.38 &  \\ 
\newcite{zhang-2013}      &&    & 93.50&\\ \hline
\newcite{koo08} $\dagger$ &&     &93.16 & \\
\newcite{carreras08} $\dagger$ &&      &93.5~~ &  \\ 
\newcite{suzuki-EtAl:2009}  $\dagger$  &&      &93.79&       \\ \hline 
joint-dynamic  & 97.42 & 92.50 &  93.50  & 15\\\hline 
\end{tabular}
}
\caption{State of the art comparison for parsing on the test set. 
Results marked with a dagger$\dagger$ are not directly comparable as additional data was used.}
\vspace{-0.4cm}
\label{table002}
\end{table}

\section{Morphological Tagging Experiments}
\label{morphology}

The joint morphology and syntactic inference requires the selection of morphological attributes (case, number, etc.) and the selection of features to predict the morphological attributes. In past work on joint morphosyntactic parsing, all morphological attributes are predicted jointly with syntactic dependencies \cite{tacl-bbjn}. However, this could lead to unnecessary complexity as only a subset of the attributes are likely to influence parsing decisions, and vice versa.

In this section we investigate whether feature selection methods can also be used to reduce the set of morphological attributes that are predicted as part of a joint system. For instance, consider a language that has the following attributes: case, gender, number, animacy. And let us say that language does not have gender agreement. Then likely only case and number will be useful in a joint system, and the gender and animacy attributes can be predicted independently. This could substantially improve the speed of the joint model -- on top of standard feature selection -- as the size of the morphosyntactic tag set will be reduced significantly.

\subsection{Data Sets}

We use the data sets listed in subsection~\ref{exsetup} for the languages that provide morphological annotation, which are German, Hungarian and Russian. 

\subsection{Main Results: Attribute Selection}

For the selection of morphological attributes (e.g. case, number, tense), we explore a simple method that departs slightly from those in Section~\ref{featselection}. In particular, we do not run greedy forward selection. Instead, we compute accuracy improvements for each attribute offline. We then independently select attributes based on these values. Our initial design was a greedy forward attribute selection, but we found experimentally that independent attribute selection worked best.

We run 10-fold cross-validation experiments on the training set where 90\% of training set is used for training and 10\% for testing. Here we simply test for each attribute independently whether its inclusion in a joint morphosyntactic parsing system increases parsing accuracy (LAS/UAS) by a statistically significant amount. If it does, then it is included. We applied cross validation to obtain more reliable results than with the development sets as some improvements where small, e.g., gender and number in German are within the range of the standard deviation results on the development set. We use parsing accuracy as we are primarily testing whether a subset of attributes can be used in place of the full set in joint morphosyntactic parsing.

Even though this method only tests an attribute's contribution independently of other attributes, we found experimentally that this was never a problem.
For instance, in German, without any morphologic attribute, we get a baseline of 89.18 LAS; when we include the attribute case,
we get 89.45 LAS; and when we include number, we get 89.32 LAS. 
When we include both case and number, we get 89.60 LAS. 

Table~\ref{attribute-selection} shows which attributes were selected.
We include 
an attribute when the cross-validation experiment shows an improvement of at least 0.1 with a statistical significance of 0.01 or better (indicated in the table by **). Some borderline cases remain such as for Russian passive where we observed an accuracy gain of 0.2 but only a low statistical significance.

\begin{table}[t]
\scriptsize
\centering
\renewcommand{\tabcolsep}{1.0pt} 
\begin{tabular}{|r|r|r|l||c|}
\hline
  \multicolumn{5}{|c|}{{\bf German}} \\\hline
 & \multicolumn{3}{|c|}{{\bf Cross Valid. Exp.}} & \\\hline
 attribute & LAS & UAS & stat. sig.  & Sel.  \\\hline
 \em{none}&  89.2 & 91.8 & -- &\\\hline
 case     &  89.5 & 91.9 & yes$^{***}$ &\X \\\hline
 gender   &  89.2& 91.8 & no          &\\\hline
 number   &  89.3& 91.9 & yes$^{***}$ &\X \\\hline
 mode     &  89.2& 91.8 & no          &\\\hline
 person   &  89.2& 91.8 & no          &\\\hline
 tense    &  89.2& 91.8 & no          &\\\hline
\multicolumn{5}{c} \ \\\hline
  \multicolumn{5}{|c|}{{\bf Hungarian}} \\\hline
 & \multicolumn{3}{|c|}{{\bf Cross Valid. Exp.}} & \\\hline
 attribute & LAS & UAS & stat. sig.  & Sel.  \\\hline
 \em{none}&  84.5& 88.3 & --          &\\\hline
 case     &  85.7& 89.0 & yes$^{****}$&\X \\\hline
 degree   &  84.6& 88.4 & yes$^{*}$   &   \\\hline
 number   &  84.7& 88.5 & yes$^{**}$  &\X \\\hline
 mode     &  84.6& 88.4 & no          & \\\hline
 person P &  84.6& 88.7 & yes$^{*}$   & \\\hline
 person   &  85.0& 88.9 & yes$^{**}$  &\X \\\hline
 subpos   &  85.4& 88.9 & yes$^{***}$ &\X  \\\hline
 tense    &  84.6& 88.4 & yes$^{*}$   & \\\hline
\end{tabular}
\begin{tabular}{|r|r|r|l||c|}
	\hline
 \multicolumn{5}{|c|}{{\bf Russian}} \\\hline
 & \multicolumn{3}{|c|}{{\bf Cross Valid. Exp.}} & \\\hline
 attribute & LAS & UAS & stat. sig.  & Sel.  \\\hline
 \em{none}&  79.4 & 88.2 &              & \\\hline
 act      &  80.4 & 89.1 & yes$^{****}$ &\X \\\hline
 anim     &  79.8 & 88.3 & yes$^{****}$ &\X \\\hline
 aspect   &  79.4 & 88.2 & no           &   \\\hline
 case     &  80.9 & 89.3 & yes$^{****}$ &\X \\\hline
 degree   &  79.4 & 88.2 & no           & \\\hline
 gender   &  80.1 & 88.6 & yes$^{****}$ &\X \\\hline
 mode     &  80.0 & 88.7 & yes$^{***}$  &\X\\\hline
 number   &  82.2 & 88.6 & yes$^{****}$ &\X\\\hline
 passive  &  79.6 & 88.2 & yes$^{*}$    &\\\hline
 tense    &  79.8 & 88.4 & yes$^{**}$   &\X\\\hline
 typo     &  79.4 & 88.1 & no           &\\\hline
\end{tabular}

\caption{Morphological attribute selection. }
\label{attribute-selection}
\end{table}

\begin{table}[t]
\small
\centering
\renewcommand{\tabcolsep}{1.0pt} 
\scalebox{1.0}{ 
\begin{tabular}{|r|r|r|r|r||l|c|c|c|c|c|c|c|c|c|c|c|c|c|c|c|c|c|c|c|c|c|c}
\hline
          &\multicolumn{4}{|c|}{{\bf Pipeline}}&\multicolumn{4}{|c|}{{\bf Joint}} \\\hline
          & MOR   & LAS   & UAS   & \#       & MOR   & LAS   & UAS    & \# \\\hline
  \multicolumn{9}{|c|}{{\bf German}} \\\hline
none      &  93.07 & 91.69 & 93.66  & 83     & 94.21& 91.69 & 93.65 &  83 \\\hline 
static    &  92.65 & 91.70 & 93.75  & 44     &\bf 94.17 & 91.56 & 93.54 &\bf  27 \\\hline 
dynamic      &  92.89 & 91.61 & 93.62  & 29     & 94.01 & 91.72 & 93.72 &  32 \\\hline 
  \multicolumn{9}{|c|}{{\bf Hungarian}} \\\hline
none      &  97.10 & 88.00 & 90.49 & 83       & 97.22 & 88.02 & 90.51 & 83\\\hline 
static    &  97.11 & 87.98 & 90.43 & 24       &\bf 97.23 & 88.06 & 90.49 & 18 \\\hline 
dynamic    &  96.90 & 87.92 & 90.38 & 15       & 97.22 & 87.97 & 90.43 &\bf 12 \\\hline 
  \multicolumn{9}{|c|}{{\bf Russian}} \\\hline
none      & 93.41 & 87.36 & 92.58 & 83     & 95.78& 87.54 & 92.79  &  83 \\\hline 
static    & 95.37 & 87.50 & 92.80 & 31     & 95.39  & 87.58 & 92.85  &  21 \\\hline 
dynamic   & 95.31 & 87.53 & 92.74 & 24     &\bf 95.74  & 87.59& 92.86&\bf 16 \\\hline 
\end{tabular}
}
\caption{Morphological and syntactic accuracy scores on the dev set without feature selection (none), with greedy forward feature selection (static) and
with  Minimum Redundancy Maximum Relevance (dynamic). 
}
\label{table:morph-results-dev}
\end{table}

\begin{table}[t]
\small
\centering
\renewcommand{\tabcolsep}{1.0pt} 
\scalebox{1.0}{ 
\begin{tabular}{|r|r|l|l|l|l|l|l|l|l|l|l|l|l|r|r|r|r|r|r|r|c|c|c|c}
\hline
System    &            \multicolumn{4}{|l|}{} \\\hline
          &            POS   & MOR & LAS   & UAS    \\\hline
  \multicolumn{5}{|c|}{{\bf German}} \\\hline
\newcite{seeker13}       &       &               & 91.50       &  93.48 \\
joint-static-static      & 97.97 & 94.20         & 91.88       &  93.81    \\\hline 
  \multicolumn{5}{|c|}{{\bf Hungarian}} \\\hline
\newcite{farkas12}       &       &       & 87.2 & 90.1   \\ 
\newcite{tacl-bbjn}      & 97.80 & 96.4$\dagger$ & 88.9  & 91.3 \\
joint-static-static      & 97.85 & 96.97         & 88.85 & 91.32   \\\hline 
  \multicolumn{5}{|c|}{{\bf Russian}} \\\hline
\newcite{boguslavsky11}  &       &               & 86.0 & 90.0 \\
\newcite{tacl-bbjn}      & 98.50 & 94.4$\dagger$ & 87.6 & 92.8 \\
joint-dynamic-dynamic    & 98.90 & 95.62 & 87.86 & 92.95  \\\hline 
\end{tabular}
}
\caption{State of the art comparison on the test set. 
Results marked with a dagger$\dagger$ are not comparable since they use different morphological attribute bundels.}
\label{table003}
\end{table}

\subsection{Main Results: Feature Selection}

Having fixed the set of attributes to be predicted jointly with the parser, we can turn our attention to optimizing the feature sets for morphosyntactic tagging.
To this end, we again consider greedy forward selection with the static and dynamic strategies. Table~\ref{table:templates} shows the selected features for the different languages where the grey boxes again mean  that the feature was selected. Table~\ref{table:morph-results-dev} shows the performance on the development set. For German, the full template set performs best but only 0.04 better than static selection which performs nearly as well while reducing the template set by 68\%. For Hungarian, all sets perform similarly while dynamic selection needs 86\% less features. The top performing feature set for Russian is dynamic selection in a joint system which needs 81\% less features. We observe again that dynamic selection tends to select less feature templates compared to static selection, but here both the full set of features and the set selected by static selection appear to have better accuracy on average. 

The feature selection methods obtain significant speed-ups for the joint system. On the development sets
we observed a speedup from 0.015 to 0.003 sec/sentence for Hungarian, from 0.014 to 0.004 sec/sentence for German, and from 0.015 to 0.006 sec/sentence for Russian. This represents a reduction in running time between 50 and 80\%.

Table~\ref{table003} compares our system to other state-of-the-art morphosyntactic parsers. We can see that on average the accuracies of our attribute/feature selection models are competitive or above the state-of-the art. The key result is that state of the art accuracy can be achieved with much leaner and faster models.

\section{Conclusions}
\label{sec:conclusion}


There are several methodological lessons to learn from this paper. First, feature selection is generally useful as it leads to fewer features and faster tagging while maintaining state-of-the-art results. Second, feature selection is even more effective for joint tagging-parsing, where it leads to even better results and smaller feature sets. In some cases, the number of feature templates is reduced by up to 80\% with a correponding reduction in running time. Third, dynamic feature selection strategies \cite{Hanchuan2005} lead to more compact models than static feature selection, without significantly impacting accuracy. Finally, similar methods can be applied to morphological attribute selection leading to even leaner and faster models.




\section*{Acknowledgement}
Miguel Ballesteros is supported by the European Commission under the contract numbers FP7-ICT-610411 (project MULTISENSOR) and H2020-RIA-645012 (project KRISTINA)





\bibliographystyle{acl}
\bibliography{xample,main3} 



\end{document}